\documentclass{article}

\usepackage{arxiv}

\usepackage[utf8]{inputenc} 
\usepackage[T1]{fontenc} 
\usepackage{hyperref} 
\usepackage{url} 
\usepackage{booktabs} 
\usepackage{amsfonts} 
\usepackage{nicefrac} 
\usepackage{microtype} 
\usepackage{lipsum} 
\usepackage{graphicx}
\usepackage{natbib}
\usepackage{doi}
\usepackage{algorithm}%
\usepackage{algorithmicx}%
\usepackage{algpseudocode}%
\usepackage{listings}%
\usepackage{amsthm}%
\usepackage{mathrsfs}%
\usepackage{amsmath}
\usepackage{booktabs}
\usepackage{tikz}
\usetikzlibrary{arrows.meta,positioning,fit}
\usetikzlibrary{calc}
\usepackage[table]{xcolor}
\usetikzlibrary{positioning}
\usepackage{tabularx}
\usepackage{multirow}

\usepackage{tabularx}

\usepackage{colortbl}
\usepackage[table]{xcolor} 
\definecolor{lightorange}{rgb}{1.0, 0.9, 0.4}

\title{Towards Automatic Heart Murmur Characterization from Multi-spot Phonocardiograms}

\date{} 

\author{
\textbf{Marcelo Nogueira}\thanks{Corresponding author.} \\
INESC TEC \& Faculdade de Ci\^encias, Universidade do Porto \\
\texttt{diogo.m.nogueira@inesctec.pt}
\And
\textbf{Jorge H. Oliveira} \\
LIACC, Universidade da Maia \\
\texttt{jorgefisicomat@gmail.com}
\And
\textbf{Carlos G. Ferreira} \\
INESC TEC \& Instituto Superior de Engenharia do Porto \\
\texttt{cgf@isep.ipp.pt}
\And
\textbf{Miguel T. Coimbra} \\
INESC TEC \& Faculdade de Ci\^encias, Universidade do Porto \\
\texttt{mcoimbra@fc.up.pt}
\And
\textbf{Al\'ipio M. Jorge} \\
INESC TEC \& Faculdade de Ci\^encias, Universidade do Porto \\
\texttt{amjorge@fc.up.pt}
}

\renewcommand{\shorttitle}{Towards Automatic Heart Murmur Characterization from Multi-spot Phonocardiograms}

\hypersetup{
pdftitle={A template for the arxiv style},
pdfsubject={q-bio.NC, q-bio.QM},
pdfauthor={David S.~Hippocampus, Elias D.~Striatum},
pdfkeywords={First keyword, Second keyword, More},
}

\begin{document}
\maketitle

\begin{abstract}
Cardiac auscultation traditionally detects the presence of a heart
murmur, but clinicians further characterize it by timing, shape,
pitch, quality, and grade. Most automatic phonocardiogram analysis
systems stop at binary detection and do not estimate this richer
description. In this work, we formulate automatic murmur
characterization as five patient-level classification tasks and
compare three multi-spot strategies on recordings from the aortic,
pulmonary, mitral, and tricuspid auscultation locations. First,
classical and deep-learning classifiers are trained independently at
each location, and predictions are combined via a maximum-severity
late-fusion rule. Second, a synchronous multi-input CNN jointly
processes one representative segment per location for a single task.
Third, a unified multi-input, multi-output CNN predicts all five
attributes from one shared multi-spot representation. Using 735
patients from the CirCor DigiScope dataset with complete four-spot
recordings, the best late-fusion configurations reach an unweighted
mean sensitivity between 85.56\% and 93.81\% across the five tasks,
with class-specific recalls between 80.00\% and 100.00\% for every
relevant category. The multi-input CNN trails late fusion by 7-16
points depending on the task, while the multi-output model reaches
values between 72.59\% and 79.40\%, estimating all five attributes
from one shared representation. We further show that naively
balancing training data by adding all minority-class segments
degrades patient-level performance under the maximum-severity rule,
since it introduces non-representative, noisy segments not filtered
by the clustering-based segment selection used for evaluation. These
results indicate multi-spot murmur characterization is feasible with
both fusion-based and jointly learned architectures, outlining a path
toward automatic, structured murmur reports complementing binary
screening.
\end{abstract}


\section{Introduction}
\label{sec:intro}

Cardiovascular diseases (CVD) remain the leading cause of death
worldwide~\cite{WHO_CVD_2021}. Among CVDs, valvular heart disease alone
affected an estimated 54.8 million people with rheumatic heart disease,
13.3 million with calcific aortic valve disease, and 15.5 million with
degenerative mitral valve disease globally in 2021, with prevalence of
calcific and degenerative valve disease continuing to rise in aging and
transitioning populations~\cite{VHD_Burden_2024,Yadgir2020}. Many of
these conditions manifest acoustically as a heart murmur, making
auscultation-based screening a uniquely scalable opportunity for early
detection: cardiac auscultation continues to be the most
cost-effective, non-invasive screening procedure available in primary
care, particularly in low-resource settings where echocardiography is
not routinely accessible. A complete auscultation examination requires
listening at the four primary locations of the chest: the aortic (AV),
pulmonary (PV), tricuspid (TV), and mitral (MV) valve areas. Automatic
phonocardiogram (PCG) analysis has predominantly targeted a binary
formulation of this task: distinguishing normal from abnormal heart
sounds, or detecting the presence or absence of a murmur, an interest
reflected in large community efforts such as the George B. Moody
PhysioNet Challenge 2022~\cite{Reyna2022_Challenge}. Such systems can
support referral decisions in resource-constrained settings, but they
do not reproduce the richer clinical description that a physician
constructs during auscultation.

When a murmur is identified, clinicians further characterize it
according to when it occurs in the cardiac cycle (timing), how its
intensity evolves (shape), its dominant spectral content (pitch), its
perceived acoustic quality, and its loudness (grade). These attributes
jointly inform the differential diagnosis and the likely underlying
pathology. Automatic estimation of this structured description, rather
than a single normal/abnormal label, would move computer-aided
auscultation closer to the reasoning used in clinical practice.

The CirCor DigiScope dataset makes this richer task tractable at scale.
In addition to recordings from up to four auscultation locations per
patient, it provides expert annotations of murmur timing, shape, pitch,
quality, and grade. This combination is uncommon: most public PCG
datasets provide only normal/abnormal or murmur-present labels. The
availability of multiple recordings per patient also raises the
question of whether multi-spot information, already shown to improve
binary murmur detection, extends to the estimation of these five
clinical attributes.

This work compares three multi-spot strategies for murmur
characterization. First, classical and deep-learning classifiers are
trained independently and asynchronously at each auscultation location;
their four predictions are then combined through a clinically motivated
maximum-severity late-fusion rule. Second, a synchronous multi-input
CNN jointly processes one representative segment from each location for
a single target task. Third, a unified multi-input, multi-output CNN
predicts all five attributes simultaneously from one shared multi-spot
representation, which is a necessary step toward an automatic,
structured murmur report.

The contributions of this work are:
\begin{itemize}
\item a formulation of automatic murmur characterization as five
clinically motivated patient-level classification tasks: timing,
shape, pitch, grade, and quality;
\item a comparative evaluation of asynchronous spot-specific
classifiers with late fusion, a synchronous multi-input CNN, and a
unified multi-input, multi-output CNN, using verified class-specific
confusion matrices for every task;
\item an analysis of the impact of severe class imbalance and of
training-set balancing strategies on patient-level, multi-spot
aggregation; and
\item evidence that a single shared multi-task model can estimate
all five attributes simultaneously, providing a foundation for
future automatic murmur reporting systems.
\end{itemize}

The rest of the paper is organized as follows. Section~\ref{sec:relwork}
reviews related work on murmur detection and characterization.
Section~\ref{sec:dataset} describes the dataset and labels.
Section~\ref{sec:methodology} details the pre-processing, feature
extraction, and the three model families. Section~\ref{sec:setup}
describes the experimental protocol and evaluation metrics.
Section~\ref{sec:results} presents and discusses the results,
Section~\ref{sec:limitations} discusses the study's limitations, and
Section~\ref{sec:conclusion} concludes the paper.

\section{Related Work}
\label{sec:relwork}

Most automatic PCG analysis systems formulate auscultation as a binary
screening task: distinguishing normal from abnormal recordings or
detecting the presence of a murmur. This formulation has supported the
development of robust screening methods, but it does not reproduce the
clinical description generated during cardiac auscultation. For
example, Parvaneh et al.~\cite{Parvaneh2022} used an ensemble of
deep-learning classifiers for recordings from multiple locations,
combining location-level predictions at the decision stage. Lee et
al.~\cite{Lee2022} transformed PCG recordings into log-Mel spectrogram,
short-time Fourier transform, and constant-Q transform representations,
then evaluated CNN and Light CNN models for murmur detection. Ding et
al.~\cite{Ding2022} combined CNN-derived Mel-spectrogram features with
physician-inspired features based on the S1, systole, S2, and diastole
phases of the cardiac cycle. These studies demonstrate the value of
automatic murmur detection and multi-spot information for screening,
but their final output remains primarily a normal/abnormal or
murmur-present/murmur-absent decision.

The importance of reliable murmur screening has also motivated
large-scale community benchmarking efforts. The George B. Moody
PhysioNet Challenge 2022 released the CirCor DigiScope
dataset~\cite{oliveira2022circor} and invited more than 40 international
teams to develop murmur detection algorithms evaluated on a shared,
multi-spot pediatric cohort~\cite{Reyna2022_Challenge}. This effort
established multi-spot PCG analysis as a benchmarked research problem
and produced strong binary detection baselines, but, consistent with
the studies discussed above, the Challenge's official scoring targets
remained murmur presence and clinical outcome rather than the detailed
murmur description considered in this work.

Multi-spot synchronous analysis of heart sounds was also investigated
by Nogueira et al.~\cite{Nogueira_SegmentSelection_2026}, who proposed
a clustering-based algorithm to select the most representative
three-second segment from each of the four auscultation locations and
showed that feeding these representative segments into a synchronous
multi-input CNN improves binary murmur/pathology detection over both
single-channel and asynchronous multi-channel baselines, achieving an
overall accuracy of 96.5\% on the same 735-patient CirCor DigiScope
cohort used in this work. The present work adopts the same
representative-segment selection strategy and extends the underlying
synchronous multi-input architecture from binary detection to the five
murmur characterization tasks described below.

Moving from detection to characterization is a comparatively recent
research direction. Elola et al.~\cite{Elola2023} addressed
patient-level murmur grading from CirCor DigiScope PCGs, classifying
each recording as absent, soft, or loud using an ensemble of 15
convolutional residual networks with squeeze-and-excitation attention,
operating on Mel spectrogram windows. Predictions from the available
auscultation locations were aggregated by favoring the highest
estimated grade. On a held-out test set of 1538 recordings from 442
patients, their method obtained an unweighted mean sensitivity of
80.4\%; class-specific sensitivities were 90.7\%, 75.8\%, and 84.2\%
for absent, soft, and loud murmurs, respectively. The comparatively
lower performance for soft murmurs illustrates the difficulty of
intermediate and clinically ambiguous grades.

Xu et al.~\cite{Xu2024} proposed an adaptable heterogeneous-modality
multi-task framework for murmur detection, grading, and cardiac risk
analysis. Their hierarchical multi-task module was designed to reduce
negative transfer between detection and grading tasks, while a
heterogeneous-modality adaptation module fused unstructured PCG
representations with structured demographic information. The work also
incorporated random-mask-based importance evaluation and SHAP
visualizations to identify informative murmur segments, illustrating
that characterization can be combined with risk modeling and
interpretability. However, the work primarily focuses on grade and
risk rather than jointly estimating the complete set of attributes
used in clinical murmur descriptions.

Quality has been studied separately by Wu et al.~\cite{Wu_2024_quality}, who
classified murmurs as harsh or blowing using 1266 segments from 164
CirCor patients. Their CNN--GRU architecture combines channel attention
at the spectrogram feature-extraction stage with a patient-level
feature attention module that weights the contribution of different
segments. Five-fold cross-validation yielded an accuracy of 73.6\%,
with F1-scores of 76.8\% for harsh murmurs and 67.8\% for blowing
murmurs. This result highlights the difficulty of automatically
estimating perceptual attributes even when murmur presence is already
known.

Together, these studies motivate three choices in the present work.
First, evaluation should go beyond raw accuracy, since clinically
important attribute classes are substantially underrepresented. Second,
the multi-spot nature of clinical auscultation should be represented
explicitly instead of treating recordings from different anatomical
locations as interchangeable independent samples. Third, the related
attributes may benefit from a shared representation, since they arise
from the same acoustic event and jointly constitute the clinical
description of a murmur. The proposed multi-input, multi-output
architecture is designed around these three considerations.

\section{Dataset and Labels}
\label{sec:dataset}

The experiments use the CirCor DigiScope dataset. The present analysis
is restricted to the 735 patients for whom recordings from all four
primary auscultation locations are available: aortic valve (AV),
pulmonary valve (PV), mitral valve (MV), and tricuspid valve (TV). This
criterion follows the clinical objective of reproducing a complete
multi-spot auscultation assessment and is also required by the
synchronous multi-input models, which require one signal from each
location.

One representative three-second segment is selected from each
available recording using the clustering-based strategy proposed by
Nogueira et al.~\cite{Nogueira_SegmentSelection_2026}. Briefly, segments
extracted from a recording are grouped using $K$-means clustering, with
the number of clusters selected through silhouette analysis. The
largest cluster is retained and the segment closest to its centroid is
selected as the representative segment. This procedure aims to reduce
the contribution of atypical or noisy portions of an ambulatory
recording while retaining one segment per location and per patient.
Nogueira et al.~\cite{Nogueira_SegmentSelection_2026} showed that this
selection strategy yields a statistically significant improvement over
random segment selection ($p=0.003$) for synchronous multi-input CNN
classification on the same four-spot patient cohort used in this work.

The target labels are patient-level descriptions derived from the
expert annotations. Five tasks are considered:
\begin{itemize}
\item \textbf{Timing}: normal, early-systolic, mid-systolic, or
holo-systolic;
\item \textbf{Shape}: normal, decrescendo, diamond, or plateau;
\item \textbf{Pitch}: normal, low, medium, or high;
\item \textbf{Grading}: normal, soft (Levine grades I--II/VI), or
loud (grades III--IV/VI); and
\item \textbf{Quality}: normal, blowing, or harsh.
\end{itemize}

Table~\ref{tab:label_distribution} summarizes the class distributions.
Each task is highly imbalanced: normal recordings represent
approximately 85\% of the selected patient population, whereas the
rarest class, decrescendo shape, includes only 13 patients. A trivial
classifier that always predicts the normal class would obtain an
accuracy of approximately 85\%, while failing completely to identify
any murmur characteristic. Consequently, accuracy is reported only as a
secondary measure; the primary evaluation metric is unweighted mean
sensitivity (UMS), defined in Section~\ref{sec:metrics}.

\begin{table}[htbp]
\centering
\small
\renewcommand{\arraystretch}{1.15}
\begin{tabular}{llr}
\toprule
\textbf{Task} & \textbf{Class} & \textbf{Patients} \\
\midrule
Timing & Normal & 625 \\
& Early-systolic & 27 \\
& Mid-systolic & 18 \\
& Holo-systolic & 64 \\
\midrule
Shape & Normal & 625 \\
& Decrescendo & 13 \\
& Diamond & 37 \\
& Plateau & 56 \\
\midrule
Pitch & Normal & 625 \\
& Low & 20 \\
& Medium & 43 \\
& High & 47 \\
\midrule
Grading & Normal & 625 \\
& Soft (I--II/VI) & 40 \\
& Loud (III--IV/VI) & 70 \\
\midrule
Quality & Normal & 625 \\
& Blowing & 35 \\
& Harsh & 74 \\
\bottomrule
\end{tabular}
\caption{Patient-level class distributions for the five murmur
characterization tasks.}
\label{tab:label_distribution}
\end{table}

\section{Methodology}
\label{sec:methodology}

\subsection{Pre-processing and MFCC Features}
\label{sec:features}

Each PCG recording is band-pass filtered to suppress low-frequency
motion artifacts and high-frequency environmental noise, and its
amplitude is normalized to the interval $[-1,1]$. Recordings are
divided into three-second segments aligned to the beginning of S1. For
each recording, one representative segment is selected through the
clustering-based procedure described in Section~\ref{sec:dataset}.

Mel-frequency cepstral coefficients (MFCCs) are extracted from each
representative segment using a 25\,ms analysis window and a 10\,ms hop
size, yielding 300 time frames per three-second segment. The number of
MFCC coefficients is treated as a hyperparameter. Four feature
dimensionalities, 2, 3, 4, and 6 coefficients per frame, are evaluated
for every model family and clinical task. This choice reflects the
trade-off between preserving discriminative spectral information and
controlling the dimensionality and complexity of the feature
representation. The selected dimensionalities were obtained through an
author-guided combination of manual exploration and model-performance
comparison, rather than through an exhaustive automated hyperparameter
search. As shown in Section~\ref{sec:results}, the best-performing
dimensionality is not the same across tasks or model families.

\begin{figure}[htbp]
\centering
\includegraphics[width=0.95\textwidth]{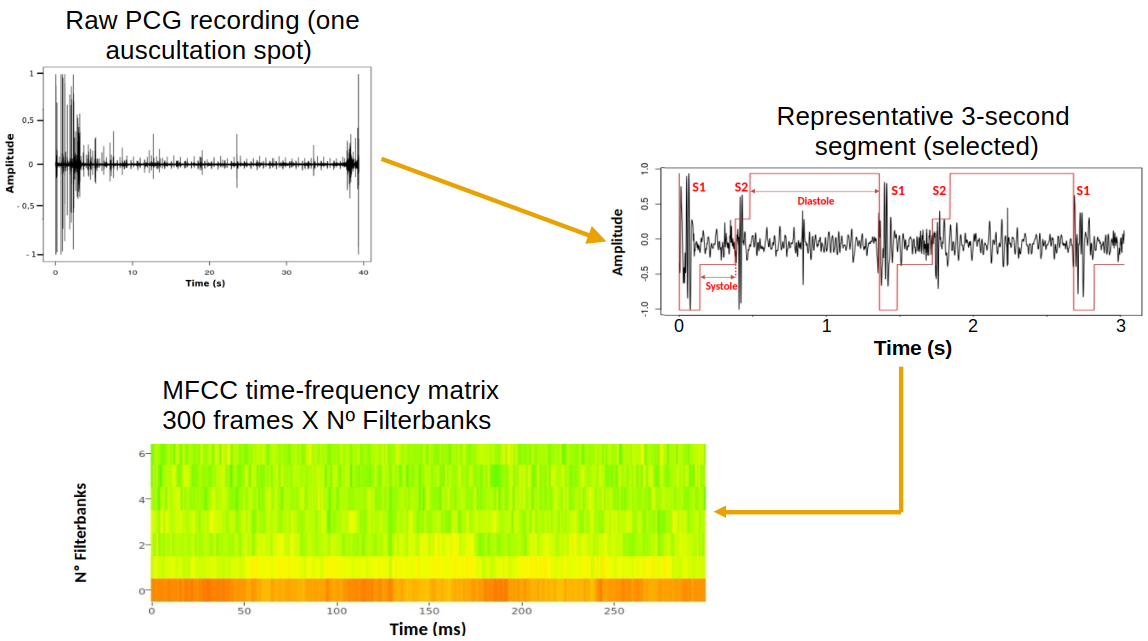}
\caption{Illustration of the pre-processing pipeline for a single
auscultation location. A raw PCG recording (top left) is segmented
into three-second windows aligned to S1, and the clustering-based
representative segment is selected (top right), with the S1, systole,
S2, and diastole intervals of one cardiac cycle highlighted. MFCCs are
then extracted from the representative segment, producing a
300-frame time-frequency matrix (bottom) that is used as model input.}
\label{fig:pcg_pipeline}
\end{figure}

\subsection{Spot-specific Models and Asynchronous Late Fusion}
\label{sec:latefusion}

The first family of experiments trains separate classifiers for each
anatomical location. Each model receives the MFCC representation from
one representative segment only and produces a spot-specific prediction
for one clinical target. Support vector machines (SVM), random forests
(RF), XGBoost, decision trees (DT), and single-input CNNs are evaluated
for each target task, each MFCC dimensionality, and each of the AV, MV,
PV, and TV locations.

The classical models were implemented in R. SVMs used a radial basis
function kernel with cost $C=1$ and $\gamma=0.001$. RF models comprised
400 trees, with $m_{try}=65$ candidate variables considered at each
split and a maximum of 50 terminal nodes. DT models were trained with
the \texttt{rpart} implementation, using \texttt{minsplit}=2,
\texttt{minbucket}=1, and maximum depth 30. XGBoost used the
\texttt{gbtree} booster with the multi-class probability objective
(\texttt{multi:softprob}), multi-class log-loss as the optimization
metric, and up to 100 boosting rounds. A stratified five-fold internal
cross-validation procedure with early stopping after 20 rounds without
improvement was used during parameter exploration. The final
hyperparameter configurations resulted from a combination of this
optimization procedure and author-guided manual exploration aimed at
maximizing validation performance while avoiding unnecessarily complex
models.

The spot-specific classifiers are trained asynchronously and
independently: no model sees recordings from more than one auscultation
location during training. Patient-level multi-spot results are then
obtained through rule-based late fusion. The four independently
generated spot predictions are combined according to a clinically
motivated maximum-severity criterion: if the predicted attribute
differs across locations, the patient is assigned the most severe
predicted class. This mirrors the clinical principle that a relevant
murmur observed at any location should not be ignored because other
locations appear normal. No additional multi-spot model is trained in
these late-fusion experiments; the patient-level decision is a
deterministic aggregation of the four independently trained
spot-specific classifiers.

\subsection{Synchronous Multi-input CNN}
\label{sec:multiinput}

A synchronous multi-input CNN is evaluated as an alternative to
asynchronous spot-specific classification followed by late fusion. The
model receives one MFCC feature matrix from each of the four
auscultation locations simultaneously. Each input is processed by a
dedicated branch with identical architecture: a 2D convolution with 64
filters of size $5\times1$ and ReLU activation, $3\times1$ max pooling,
a second 2D convolution with 32 filters of size $5\times2$ and ReLU
activation, a second $3\times1$ max-pooling operation, dropout with
rate 0.25, and flattening. The four resulting vectors are concatenated
into a joint multi-spot representation.

The fused representation is processed by dense layers of 5096, 1024,
and 128 ReLU units, followed by dropout with rate 0.4 and a
task-specific softmax output layer. A separate multi-input CNN is
trained for each murmur attribute. Models are optimized with Adam
(initial learning rate $10^{-4}$ and decay $10^{-6}$) using categorical
cross-entropy. Training uses shuffled mini-batches, a 20\% validation
split, and a checkpointing strategy that retains the model with the
lowest validation loss for final testing. The model architecture and
hyperparameters were selected through author-guided manual exploration
of alternative configurations, balancing validation performance and
model complexity.

\subsection{Multi-input Multi-output CNN}
\label{sec:multioutput}

The proposed unified model extends the synchronous multi-input CNN by
predicting all five murmur attributes from one shared multi-spot
representation. The architecture has four inputs, corresponding to AV,
MV, PV, and TV. Each branch consists of two convolutional blocks. Each
block applies a 2D convolution with ReLU activation, batch
normalization, $5\times1$ max pooling, and dropout with rate 0.3. The
first and second convolutional layers use 32 and 64 filters,
respectively. The resulting four feature maps are flattened and
concatenated.

The shared concatenated representation feeds five task-specific output
heads, corresponding respectively to timing, shape, pitch, grading, and
quality. Each head consists of a dense layer with 1024 ReLU units and
$L_2$ regularization ($\lambda=0.001$), a dropout layer with rate 0.5,
a second dense layer with 256 ReLU units, and a final softmax layer
with the number of units equal to the number of classes in the
corresponding task. The architecture, the number of convolutional
filters and dense units, the regularization settings, and the training
hyperparameters were obtained through an author-guided manual
exploration of alternative configurations, aimed at optimizing
validation performance while maintaining manageable model complexity.

The model is trained with Adam optimization, with an initial learning
rate of $10^{-4}$ and decay of $10^{-6}$. Every output is optimized
with categorical cross-entropy, and equal weights are assigned
implicitly to the five tasks:
\begin{equation}
\mathcal{L}_{\mathrm{total}} =
\mathcal{L}_{\mathrm{timing}} +
\mathcal{L}_{\mathrm{shape}} +
\mathcal{L}_{\mathrm{pitch}} +
\mathcal{L}_{\mathrm{grading}} +
\mathcal{L}_{\mathrm{quality}}.
\label{eq:multitask_loss}
\end{equation}
Training uses shuffled mini-batches and a 20\% validation split; the
checkpoint with the lowest aggregate validation loss is retained for
evaluation.

\begin{figure}[htbp]
\centering
\footnotesize
\begin{tikzpicture}[
every node/.style={align=center},
input/.style={draw=blue!65!black, fill=blue!7, rounded corners=2pt,
minimum width=1.35cm, minimum height=0.55cm, inner sep=2pt},
method/.style={draw=orange!80!black, fill=orange!13, rounded corners=2pt,
minimum width=5.55cm, minimum height=0.72cm, inner sep=4pt},
process/.style={draw=black!65, fill=black!5, rounded corners=2pt,
minimum width=5.55cm, minimum height=0.66cm, inner sep=4pt},
result/.style={draw=green!50!black, fill=green!10, rounded corners=2pt,
minimum width=1.35cm, minimum height=0.55cm, inner sep=2pt},
bus/.style={minimum width=3.2cm, minimum height=0pt, inner sep=0pt},
title/.style={font=\bfseries\footnotesize},
arrow/.style={-{Stealth[length=1.7mm]}, line width=0.55pt}
]
\node[title] (titlea) {A. Asynchronous spot-specific classification and late fusion};
\node[input, below=3mm of titlea, xshift=-2.15cm] (aav) {AV};
\node[input, right=1.5mm of aav] (amv) {MV};
\node[input, right=1.5mm of amv] (apv) {PV};
\node[input, right=1.5mm of apv] (atv) {TV};
\node[method, below=12mm of amv, xshift=0.75cm] (modela) {Independent models: SVM, RF, XGBoost, DT, or CNN\\trained separately at each auscultation location};
\node[process, below=9mm of modela] (fusea) {Four spot-specific predictions aggregated by the maximum-severity rule};
\node[result, below=9mm of fusea, minimum width=2.90cm] (outa) {One patient-level attribute};
\draw[arrow] (aav.south) -- (aav.south |- modela.north);
\draw[arrow] (amv.south) -- (amv.south |- modela.north);
\draw[arrow] (apv.south) -- (apv.south |- modela.north);
\draw[arrow] (atv.south) -- (atv.south |- modela.north);
\draw[arrow] (modela) -- (fusea);
\draw[arrow] (fusea) -- (outa);

\node[title, below=12mm of outa] (titleb) {B. Synchronous multi-input CNN};
\node[input, below=3mm of titleb, xshift=-2.15cm] (bav) {AV};
\node[input, right=1.5mm of bav] (bmv) {MV};
\node[input, right=1.5mm of bmv] (bpv) {PV};
\node[input, right=1.5mm of bpv] (btv) {TV};
\node[bus, below=6mm of bmv, xshift=0.75cm] (bbus) {};
\node[method, below=6mm of bbus] (modelb) {Four input branches processed jointly, concatenated,\\and classified by one task-specific CNN};
\node[result, below=9mm of modelb, minimum width=2.90cm] (outb) {One patient-level attribute};
\draw[arrow] (bav.south) -- ++(0,-3mm) -| (bbus);
\draw[arrow] (bmv.south) -- ++(0,-3mm) -| (bbus);
\draw[arrow] (bpv.south) -- ++(0,-3mm) -| (bbus);
\draw[arrow] (btv.south) -- ++(0,-3mm) -| (bbus);
\draw[arrow] (bbus) -- (modelb.north);
\draw[arrow] (modelb) -- (outb);

\node[title, below=12mm of outb] (titlec) {C. Synchronous multi-input multi-output CNN};
\node[input, below=3mm of titlec, xshift=-2.15cm] (cav) {AV};
\node[input, right=1.5mm of cav] (cmv) {MV};
\node[input, right=1.5mm of cmv] (cpv) {PV};
\node[input, right=1.5mm of cpv] (ctv) {TV};
\node[bus, below=6mm of cmv, xshift=0.75cm] (cbus) {};
\node[method, below=6mm of cbus] (modelc) {Four input branches form one shared multi-spot representation};
\node[process, below=9mm of modelc] (headsc) {Five task-specific output heads};
\node[result, below=9mm of headsc, xshift=-3.10cm] (timing) {Timing};
\node[result, right=1.5mm of timing] (shape) {Shape};
\node[result, right=1.5mm of shape] (pitch) {Pitch};
\node[result, right=1.5mm of pitch] (grading) {Grading};
\node[result, right=1.5mm of grading] (quality) {Quality};
\draw[arrow] (cav.south) -- ++(0,-3mm) -| (cbus);
\draw[arrow] (cmv.south) -- ++(0,-3mm) -| (cbus);
\draw[arrow] (cpv.south) -- ++(0,-3mm) -| (cbus);
\draw[arrow] (ctv.south) -- ++(0,-3mm) -| (cbus);
\draw[arrow] (cbus) -- (modelc.north);
\draw[arrow] (modelc) -- (headsc);
\draw[arrow] (headsc.south) -- ++(0,-3mm) -| (timing.north);
\draw[arrow] (headsc.south) -- ++(0,-3mm) -| (shape.north);
\draw[arrow] (headsc.south) -- ++(0,-3mm) -| (pitch.north);
\draw[arrow] (headsc.south) -- ++(0,-3mm) -| (grading.north);
\draw[arrow] (headsc.south) -- ++(0,-3mm) -| (quality.north);
\end{tikzpicture}
\caption{Summary of the three multi-spot strategies evaluated in this
work. In each approach, one representative segment from the AV, MV,
PV, and TV locations is used. In Approach A, each location is modeled
independently and asynchronously, and the four resulting predictions
are combined through late fusion. Approach B learns one joint
representation for a single target task. Approach C shares one learned
multi-spot representation across five clinical attribute heads.}
\label{fig:approaches}
\end{figure}
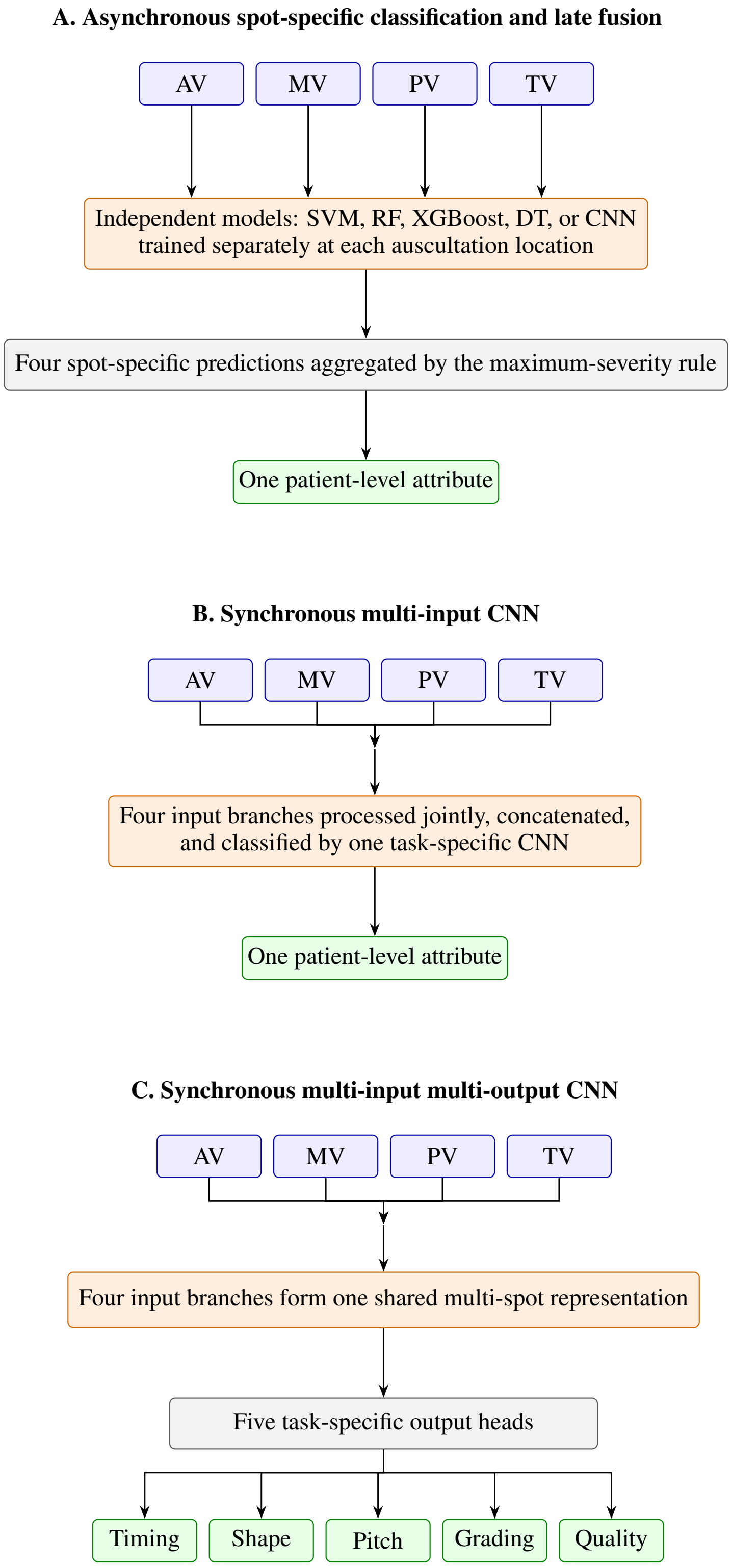

\section{Experimental Setup and Evaluation Metrics}
\label{sec:setup}

All experiments use patient-level five-fold stratified cross-validation.
Patients, and therefore all recordings belonging to a patient, are
assigned exclusively to either the training or test partition of a
given fold. This patient-independent protocol prevents information
leakage between the training and test sets. The natural class
distribution is retained in the test folds. In selected experiments,
additional three-second segments from minority-class patients not
present in the test set are included in the training data to
investigate the effect of class balancing. As discussed in
Section~\ref{sec:balancing}, this strategy has different effects for
spot-specific models and late patient-level fusion.

\subsection{Evaluation Metrics}
\label{sec:metrics}

For every fold, predictions are compared with the true patient-level
labels through a confusion matrix. Let $C_{ij}$ denote the number of
samples whose true class is $i$ and predicted class is $j$, and let $K$
denote the number of classes. Class-specific precision, recall, and
F1-score are computed as
\begin{equation}
\mathrm{Precision}_i = \frac{C_{ii}}{\sum_j C_{ji}}, \qquad
\mathrm{Recall}_i = \frac{C_{ii}}{\sum_j C_{ij}}, \qquad
\mathrm{F1}_i = \frac{2\,\mathrm{Precision}_i\,\mathrm{Recall}_i}
{\mathrm{Precision}_i+\mathrm{Recall}_i}.
\end{equation}

The primary metric is unweighted mean sensitivity (UMS), also known as
macro-average recall,
\begin{equation}
\mathrm{UMS} = \frac{1}{K}\sum_{i=1}^{K}\mathrm{Recall}_i.
\label{eq:ums}
\end{equation}
Unlike accuracy, UMS assigns equal importance to each clinical class,
regardless of its prevalence. Accuracy and the F1-score computed from
the aggregated confusion matrix are reported as complementary measures.
The displayed confusion matrices are formed after combining predictions
from all test folds.

\section{Results and Discussion}
\label{sec:results}

\subsection{General Observations}

The five characterization tasks are substantially more difficult than
binary murmur detection because all have severe class imbalance and
because the target labels represent finer distinctions between acoustic
patterns. Across timing, shape, pitch, grading, and quality, the normal
class represents approximately 85\% of the patient population. A
classifier that always predicts normal would therefore achieve an
accuracy near 85\%, while yielding an UMS equal to $1/K$, where $K$ is
the number of classes. The results below consequently prioritize UMS
and class-specific confusion matrices over accuracy alone.

A consistent pattern emerges across tasks. Classical late-fusion
models, most often RF or XGBoost, achieve the highest UMS and
F1-scores when combining the four auscultation locations. The best
algorithm and MFCC dimensionality are not identical across tasks:
timing and grading favor 6 MFCC coefficients, shape favors RF with 3
MFCC coefficients, and pitch favors XGBoost with 3 MFCC coefficients,
while quality favors XGBoost with 4 MFCC coefficients. This indicates
that the optimal time-frequency resolution depends on the specific
acoustic cues that characterize each clinical attribute. The
synchronous multi-input CNN generally trades some task-specific peak
performance for a jointly learned representation of the four
locations; across all five tasks, its UMS remains 7 to 16 percentage
points below the corresponding late-fusion result. Given the scale of
the underlying clinical need described in Section~\ref{sec:intro}, even
partial automation of murmur characterization -- for example, reliably
flagging grade or timing at the level achieved by the late-fusion
models below -- could meaningfully support triage in settings where
specialist auscultation or echocardiography is not readily available.

\begin{table}[htbp]
\centering
\small
\renewcommand{\arraystretch}{1.15}
\setlength{\tabcolsep}{3.5pt}
\begin{tabular}{l l c r r}
\toprule
\textbf{Attribute} & \textbf{Model family} & \textbf{MFCC} & \textbf{UMS (\%)} & \textbf{F1 (\%)} \\
\midrule
\multirow{2}{*}{Timing}
& Late fusion (RF/XGBoost) & 6 & 86.52 & 92.20 \\
& Multi-input CNN & 4 & 78.26 & 71.90 \\
\addlinespace
\multirow{2}{*}{Shape}
& Late fusion (RF) & 3 & 86.47 & 91.95 \\
& Multi-input CNN & 2 & 73.86 & 78.45 \\
\addlinespace
\multirow{2}{*}{Pitch}
& Late fusion (XGBoost) & 3 & 85.56 & 91.60 \\
& Multi-input CNN & 3 & 76.11 & 77.98 \\
\addlinespace
\multirow{2}{*}{Grading}
& Late fusion (XGBoost) & 6 & 93.81 & 96.49 \\
& Multi-input CNN & 2 & 78.16 & 78.76 \\
\addlinespace
\multirow{2}{*}{Quality}
& Late fusion (XGBoost) & 4 & 86.97 & 92.02 \\
& Multi-input CNN & 3 & 80.00 & 78.88 \\
\bottomrule
\end{tabular}
\caption{Best patient-level results for each murmur-characterization
attribute, comparing the best late-fusion configuration with the best
synchronous multi-input CNN configuration. UMS is the primary metric
because all tasks are severely imbalanced.}
\label{tab:main_results}
\end{table}

\subsection{Grading}

Murmur grading distinguishes normal recordings from soft (grades
I--II/VI) and loud (grades III--IV/VI) murmurs. The best late-fusion
result is achieved by XGBoost with six MFCC coefficients, reaching an
UMS of 93.81\%, an F1-score of 96.49\%, and an accuracy of 98.64\%. The
associated aggregated confusion matrix (Table~\ref{tab:grading_confusion})
shows that the normal class is identified with 99.84\% recall, while
soft and loud murmurs each reach 90.00\% recall.

\begin{table}[htbp]
\centering
\small
\renewcommand{\arraystretch}{1.12}
\setlength{\tabcolsep}{5pt}
\begin{tabular}{lrrr|r}
\toprule
& \multicolumn{3}{c}{\textbf{Predicted}} & \\
\cmidrule(lr){2-4}
\textbf{True class} & Normal & Soft & Loud & \textbf{Recall (\%)} \\
\midrule
Normal & 624 & 0 & 1 & 99.84 \\
Soft & 4 & 36 & 0 & 90.00 \\
Loud & 7 & 0 & 63 & 90.00 \\
\bottomrule
\end{tabular}
\caption{Aggregated patient-level confusion matrix for grading: best
late-fusion model (XGBoost, 6 MFCC).}
\label{tab:grading_confusion}
\end{table}

The multi-input CNN obtains its strongest grading result with two MFCC
coefficients (UMS 78.16\%, F1-score 78.76\%, accuracy 91.56\%). Grading
shows the largest UMS gap between late fusion and the synchronous
multi-input CNN among the five tasks (15.65 percentage points),
suggesting that the binary soft/loud distinction benefits substantially
from spot-specific modeling and the maximum-severity aggregation rule.

\subsection{Timing}

For timing, the task distinguishes normal recordings from early-,
mid-, and holo-systolic murmurs. The best late-fusion result is
achieved with 6 MFCC coefficients, reaching an UMS of 86.52\%, an
F1-score of 92.20\%, and an accuracy of 97.28\%. The aggregated
confusion matrix (Table~\ref{tab:timing_confusion}) shows perfect
recall for the normal class (100.00\%), 81.48\% recall for
early-systolic, 83.33\% for mid-systolic, and 81.25\% for holo-systolic
murmurs.

\begin{table}[htbp]
\centering
\small
\renewcommand{\arraystretch}{1.12}
\setlength{\tabcolsep}{3.5pt}
\begin{tabular}{lrrrr|r}
\toprule
& \multicolumn{4}{c}{\textbf{Predicted}} & \\
\cmidrule(lr){2-5}
\textbf{True class} & Normal & Early & Mid & Holo & \textbf{Recall (\%)} \\
\midrule
Normal & 625 & 0 & 0 & 0 & 100.00 \\
Early-systolic & 5 & 22 & 0 & 0 & 81.48 \\
Mid-systolic & 3 & 0 & 15 & 0 & 83.33 \\
Holo-systolic & 12 & 0 & 0 & 52 & 81.25 \\
\bottomrule
\end{tabular}
\caption{Aggregated patient-level confusion matrix for timing: best
late-fusion configuration (RF/XGBoost, 6 MFCC).}
\label{tab:timing_confusion}
\end{table}

The synchronous multi-input CNN reaches its best result with 4 MFCC
coefficients (UMS 78.26\%, F1-score 71.90\%, accuracy 88.83\%). Its
confusion matrix shows a markedly lower recall for early-systolic
murmurs (62.96\%) compared with the late-fusion model, indicating that
this minority class benefits more from the asynchronous, spot-specific
modeling strategy than from the jointly learned multi-spot
representation.

\subsection{Shape}

Shape classification includes normal, decrescendo, diamond, and
plateau murmurs. The rarest class, decrescendo, includes only 13
patients, which makes this task particularly sensitive to individual
misclassifications. The best late-fusion result is obtained with RF
using 3 MFCC coefficients, reaching an UMS of 86.47\%, an F1-score of
91.95\%, and an accuracy of 97.13\%. The aggregated confusion matrix
(Table~\ref{tab:shape_confusion}) shows recalls of 99.84\% for normal,
80.36\% for plateau, 81.08\% for diamond, and 84.62\% for decrescendo.

\begin{table}[htbp]
\centering
\small
\renewcommand{\arraystretch}{1.12}
\setlength{\tabcolsep}{3.5pt}
\begin{tabular}{lrrrr|r}
\toprule
& \multicolumn{4}{c}{\textbf{Predicted}} & \\
\cmidrule(lr){2-5}
\textbf{True class} & Normal & Plateau & Diamond & Decres. & \textbf{Recall (\%)} \\
\midrule
Normal & 624 & 1 & 0 & 0 & 99.84 \\
Plateau & 11 & 45 & 0 & 0 & 80.36 \\
Diamond & 7 & 0 & 30 & 0 & 81.08 \\
Decrescendo & 2 & 0 & 0 & 11 & 84.62 \\
\bottomrule
\end{tabular}
\caption{Aggregated patient-level confusion matrix for shape: best
late-fusion configuration (RF, 3 MFCC).}
\label{tab:shape_confusion}
\end{table}

The multi-input CNN obtains its best result with 2 MFCC coefficients
(UMS 73.86\%, F1-score 78.45\%, accuracy 92.34\%). Its confusion
matrix shows a substantially lower recall for decrescendo shape
(61.54\% versus 84.62\% for late fusion) and for plateau (64.29\%
versus 80.36\%), confirming that the rarest and most acoustically
subtle shape classes remain harder for the jointly learned
representation to separate.

\subsection{Pitch}

Pitch is categorized as normal, low, medium, or high. The best
late-fusion result is achieved by XGBoost with 3 MFCC coefficients,
reaching an UMS of 85.56\%, an F1-score of 91.60\%, and an accuracy of
97.14\%. The corresponding confusion matrix
(Table~\ref{tab:pitch_confusion}) shows recalls of 99.84\% for normal,
80.00\% for low, 81.40\% for medium, and 80.85\% for high pitch.

\begin{table}[htbp]
\centering
\small
\renewcommand{\arraystretch}{1.12}
\setlength{\tabcolsep}{3.5pt}
\begin{tabular}{lrrrr|r}
\toprule
& \multicolumn{4}{c}{\textbf{Predicted}} & \\
\cmidrule(lr){2-5}
\textbf{True class} & Normal & Low & Medium & High & \textbf{Recall (\%)} \\
\midrule
Normal & 624 & 0 & 1 & 0 & 99.84 \\
Low & 4 & 16 & 0 & 0 & 80.00 \\
Medium & 8 & 0 & 35 & 0 & 81.40 \\
High & 9 & 0 & 0 & 38 & 80.85 \\
\bottomrule
\end{tabular}
\caption{Aggregated patient-level confusion matrix for pitch: best
late-fusion configuration (XGBoost, 3 MFCC).}
\label{tab:pitch_confusion}
\end{table}

The synchronous multi-input CNN achieves its best result with 3 MFCC
coefficients (UMS 76.11\%, F1-score 77.98\%, accuracy 91.70\%). The
confusion matrix reveals recalls of 75.00\% for low, 69.77\% for
medium, and only 63.83\% for high pitch, all noticeably lower than the
corresponding late-fusion recalls. Pitch is therefore the attribute for
which the multi-input CNN struggles most with individual minority
classes, even though its overall UMS gap to late fusion (9.45
percentage points) is not the largest among the five tasks.

\subsection{Quality}

Quality classification distinguishes normal, blowing, and harsh
murmurs. The best late-fusion result is achieved by XGBoost with 4
MFCC coefficients, reaching an UMS of 86.97\%, an F1-score of 92.02\%,
and an accuracy of 97.00\%. The confusion matrix
(Table~\ref{tab:quality_confusion}) indicates 99.84\% recall for
normal, 80.00\% for blowing, and 81.08\% for harsh murmurs.

\begin{table}[htbp]
\centering
\small
\renewcommand{\arraystretch}{1.12}
\setlength{\tabcolsep}{5pt}
\begin{tabular}{lrrr|r}
\toprule
& \multicolumn{3}{c}{\textbf{Predicted}} & \\
\cmidrule(lr){2-4}
\textbf{True class} & Normal & Blowing & Harsh & \textbf{Recall (\%)} \\
\midrule
Normal & 624 & 0 & 1 & 99.84 \\
Blowing & 7 & 28 & 0 & 80.00 \\
Harsh & 14 & 0 & 60 & 81.08 \\
\bottomrule
\end{tabular}
\caption{Aggregated patient-level confusion matrix for quality: best
late-fusion configuration (XGBoost, 4 MFCC).}
\label{tab:quality_confusion}
\end{table}

The best synchronous multi-input CNN uses 3 MFCC coefficients and
reaches an UMS of 80.00\%, an F1-score of 78.88\%, and an accuracy of
91.14\%, the smallest UMS gap to late fusion among the five attributes
(6.97 percentage points). Its confusion matrix shows recalls of
71.43\% for blowing and 74.32\% for harsh murmurs, both about 7 to 9
percentage points below the late-fusion recalls, but still indicating
that quality-relevant acoustic information is captured comparatively
well by the jointly learned multi-spot representation.

\begin{table}[htbp]
\centering
\small
\renewcommand{\arraystretch}{1.18}
\setlength{\tabcolsep}{4pt}
\begin{tabular}{l l r}
\toprule
\textbf{Attribute} & \textbf{Class-specific recall (\%), best late-fusion model} & \textbf{UMS (\%)} \\
\midrule
Timing & Normal: 100.00; Early: 81.48; Mid: 83.33; Holo: 81.25 & 86.52 \\
Shape & Normal: 99.84; Plateau: 80.36; Diamond: 81.08; Decrescendo: 84.62 & 86.47 \\
Pitch & Normal: 99.84; Low: 80.00; Medium: 81.40; High: 80.85 & 85.56 \\
Grading & Normal: 99.84; Soft: 90.00; Loud: 90.00 & 93.81 \\
Quality & Normal: 99.84; Blowing: 80.00; Harsh: 81.08 & 86.97 \\
\bottomrule
\end{tabular}
\caption{Class-specific recall for the best late-fusion configuration
of each attribute, recomputed directly from the corresponding
confusion matrices.}
\label{tab:class_recalls}
\end{table}

\subsection{Impact of Class Balancing and Segment Selection}
\label{sec:balancing}

Given the severe imbalance of the five characterization tasks,
additional experiments were performed for the grading task to
investigate the impact of a more balanced training dataset. In these
experiments, additional three-second segments from minority-class
patients were included in each training fold. Patient independence was
preserved: no segments from a patient assigned to the test fold were
added to the corresponding training fold.

The balanced-training experiments improved UMS for some individual
spot-specific SVM models. However, their effect on patient-level late
fusion was detrimental. For example, the best balanced late-fusion SVM
configuration for grading reached an UMS of 77.89\%, an F1-score of
64.90\%, and an accuracy of 79.59\%, substantially below the unbalanced
late-fusion XGBoost result (UMS 93.81\%, F1-score 96.49\%, accuracy
98.64\%).

This decline is explained by two interacting factors. First, the
late-fusion rule assigns the most severe predicted grade across the
four locations to the patient. Consequently, an increase in false
positive soft or loud predictions at any individual location
propagates directly to the final patient-level decision; a patient can
be assigned Normal only when all four locations are predicted as
Normal. Second, balancing requires the inclusion of all available
minority-class segments in the training folds. Unlike the patient-level test inputs, which are selected using the
clustering-based representative-segment procedure of Nogueira et
al.~\cite{Nogueira_SegmentSelection_2026}, the additional
minority-class training segments include recordings contaminated by
stethoscope movement, background noise, or other artifacts. These segments are not
necessarily representative of the clinical class to which they belong.
The results therefore show that conventional segment-level balancing
can be counterproductive when patient-level decisions use a
maximum-severity aggregation rule and when the added minority examples
have heterogeneous signal quality.

\subsection{Unified Multi-output Characterization}

The multi-input multi-output model provides a single architecture that
produces all five elements of a structured murmur description from
four synchronously analyzed spots. Its strongest UMS values range from
72.59\% for pitch to 79.40\% for quality, consistent with the pattern
observed for the single-task multi-input CNN, where pitch is
consistently the most difficult attribute and quality benefits most
from the joint multi-spot representation. Although these values do not
exceed the highest late-fusion results, the model has a different
objective: it learns a shared multi-spot representation and returns
timing, shape, pitch, grade, and quality simultaneously. This property
is important for future automatic reporting tools, where an integrated
clinical description is more useful than five unrelated binary or
multi-class predictions generated by independent pipelines.

The current implementation assigns equal loss weight to every task;
future work should investigate whether task-dependent weights or
uncertainty-aware multi-task optimization can better balance the
learning of tasks with different class distributions and levels of
annotation ambiguity, particularly for pitch and for the rarest
classes such as decrescendo shape.

\section{Limitations}
\label{sec:limitations}

Several limitations should be considered. First, all experiments use a
single dataset and only the 735 patients with complete recordings from
all four auscultation locations. This criterion is appropriate for
synchronous multi-input models but excludes patients with incomplete
multi-spot assessments. Second, all attribute tasks are severely
imbalanced and some clinically relevant classes, such as decrescendo
shape, have very few examples. While UMS mitigates the misleading
influence of the majority class during evaluation, larger and more
balanced external datasets are needed to establish the generalizability
of the reported results. Third, the late-fusion severity rule is
clinically motivated but may amplify spot-level false positives, as
demonstrated by the balanced-training experiments in
Section~\ref{sec:balancing}. Fourth, the best model family and MFCC
dimensionality differ across attributes, which suggests that a single
fixed feature configuration is unlikely to be optimal for every
clinical task. Finally, the multi-output model uses equally weighted
task losses and does not yet exploit potential hierarchical
constraints among attributes, such as the relationship between timing
and shape or between grade and quality.

\section{Conclusion}
\label{sec:conclusion}

This work extends automatic heart sound analysis from binary murmur
detection to clinically meaningful murmur characterization. Five
patient-level attributes---timing, shape, pitch, grade, and
quality---were estimated from four multi-spot PCG recordings of the
CirCor DigiScope dataset, comparing independently trained spot-specific
classifiers combined by late fusion, task-specific synchronous
multi-input CNNs, and a unified synchronous multi-input, multi-output
CNN.

The results show that representative segment selection, multi-spot
information, and patient-level aggregation enable strong performance
across all five tasks. The highest results were obtained by classical
late-fusion models, with UMS ranging from 85.56\% for pitch to 93.81\%
for grading. Grading achieved the strongest overall result, with a
90.00\% recall for both soft and loud murmurs, while timing, shape,
pitch, and quality all reached UMS values between 85.56\% and 86.97\%
with recalls in the 80--85\% range for every minority class. The
class-specific recall analysis, recomputed directly from the reported
confusion matrices, confirms that these results cannot be explained
solely by the predominance of normal recordings. The best-performing
algorithm and MFCC dimensionality are not uniform across attributes,
indicating that the acoustic cues underlying each clinical attribute
have different optimal time-frequency resolutions, a finding with
direct implications for the design of future unified characterization
pipelines.

The strong late-fusion results must be interpreted in light of the
clinically motivated maximum-severity rule, which favors sensitivity to
relevant findings but can amplify individual spot-level false
positives. The balanced-training experiments illustrate this concern
directly: adding all available minority-class segments improved some
spot-level classifiers but substantially degraded the grading
late-fusion result (UMS falling from 93.81\% to 77.89\%). This
observation supports the use of representative segment selection and
shows that conventional balancing strategies must be assessed at the
same patient level at which clinical decisions are made.

The synchronous multi-input CNN shows that a model can learn directly
from four synchronous locations, without relying on a hand-designed
decision rule to combine four final predictions, though at a
consistent UMS cost relative to late fusion (7 to 16 percentage points
across the five tasks). This gap is smallest for quality (6.97 points)
and largest for grading (15.65 points), suggesting that binary or
near-binary severity judgments benefit more from asynchronous,
spot-specific modeling than perceptual attributes such as quality. The
multi-input multi-output CNN further establishes that the five
clinical attributes can be estimated simultaneously from one shared
multi-spot representation, with UMS values ranging from 72.59\% to
79.40\%. Although these values do not yet exceed the strongest
specialized late-fusion models, the unified model offers a
complementary practical advantage: it generates timing, shape, pitch,
grade, and quality together, moving toward an automatic, structured
murmur report rather than a collection of independent screening
decisions.

Future research should refine shared multi-task representations,
incorporate class-aware and uncertainty-aware losses, evaluate
hierarchical relationships among murmur attributes, and validate the
approach on external populations. Particular attention should be paid
to attributes and classes that are currently harder for joint
multi-spot models, namely pitch and the rarer shape and timing
categories, since these are precisely the cases where an integrated,
report-oriented system would need to either improve its sensitivity or
communicate its uncertainty to the clinician.

\vspace{0.9em}

\textbf{Data Availability} The public CirCor DigiScope Dataset used in this study are available on their original paper and website. The code used to implement the segment selection algorithm and the multi-input CNN model described in this work is available from the corresponding author upon reasonable request.

\section*{Declarations}

\textbf{Competing interests:} The authors declare no competing interests.

\vspace{0.9em}

\textbf{Ethics approval} The research conducted for this paper adheres to ethical principles and guidelines concerning the utilization of publicly available datasets. The datasets employed in this study, The CirCor DigiScope Dataset are publicly accessible resources without individual identifiers, thus obviating the need for specific consent from individuals.

\bibliographystyle{unsrtnat}
\bibliography{references}

\end{document}